\documentclass{article}
\usepackage{ijcai19}

\usepackage{times}
\usepackage{soul}
\usepackage{url}
\usepackage[hidelinks]{hyperref}
\usepackage[utf8]{inputenc}
\usepackage[small]{caption}
\usepackage{graphicx}
\usepackage{amsmath}
\usepackage{booktabs}
\usepackage{algorithm}
\usepackage{algorithmic}
\usepackage{multirow}
\usepackage{enumitem}
\usepackage{amssymb} 
\usepackage[table,dvipsnames]{xcolor}

\usepackage{newfloat}
\usepackage{listings}

\title{Draw Like an Artist: Complex Scene Generation with Diffusion Model via Composition, Painting, and Retouching}

\author{
Minghao Liu$^1$\and
Le Zhang$^2$\and
Yingjie Tian$^1$\footnotemark[1]\and
Xiaochao Qu$^2$\and
Luoqi Liu$^2$\and
Ting Liu$^2$\footnotemark[1]\\
\affiliations
$^1$University of Chinese Academy of Sciences, Beijing, China \\
$^2$MT Lab, Meitu Inc., Beijing, China\\
}

\begin{document}

\maketitle
\footnotetext[1]{Corresponding authors.}
\begin{abstract}
Recent advances in text-to-image diffusion models have demonstrated impressive capabilities in image quality. However, complex scene generation remains relatively unexplored, and even the definition of ‘complex scene’ itself remains unclear. In this paper, we address this gap by providing a precise definition of complex scenes and introducing a set of \textbf{C}omplex \textbf{D}ecomposition \textbf{C}riteria (\textbf{CDC}) based on this definition. Inspired by the artist's painting process, we propose a training-free diffusion framework called \textbf{C}omple\textbf{x} \textbf{D}iffusion (\textbf{CxD}), which divides the process into three stages: composition, painting and retouching. Our method leverages the powerful chain-of-thought capabilities of large language models (LLMs) to decompose complex prompts based on CDC and to manage composition and layout. We then develop an attention modulation method that guides simple prompts to specific regions to complete the complex scene painting. Finally, we inject the detailed output of the LLM into a retouching model to enhance the image details, thus implementing the retouching stage. Extensive experiments demonstrate that our method outperforms previous SOTA approaches, significantly improving the generation of high-quality, semantically consistent, and visually diverse images for complex scenes, even with intricate prompts.
\end{abstract}

\section{Introduction}

\begin{figure}
    \centering
    \includegraphics[width=\linewidth]{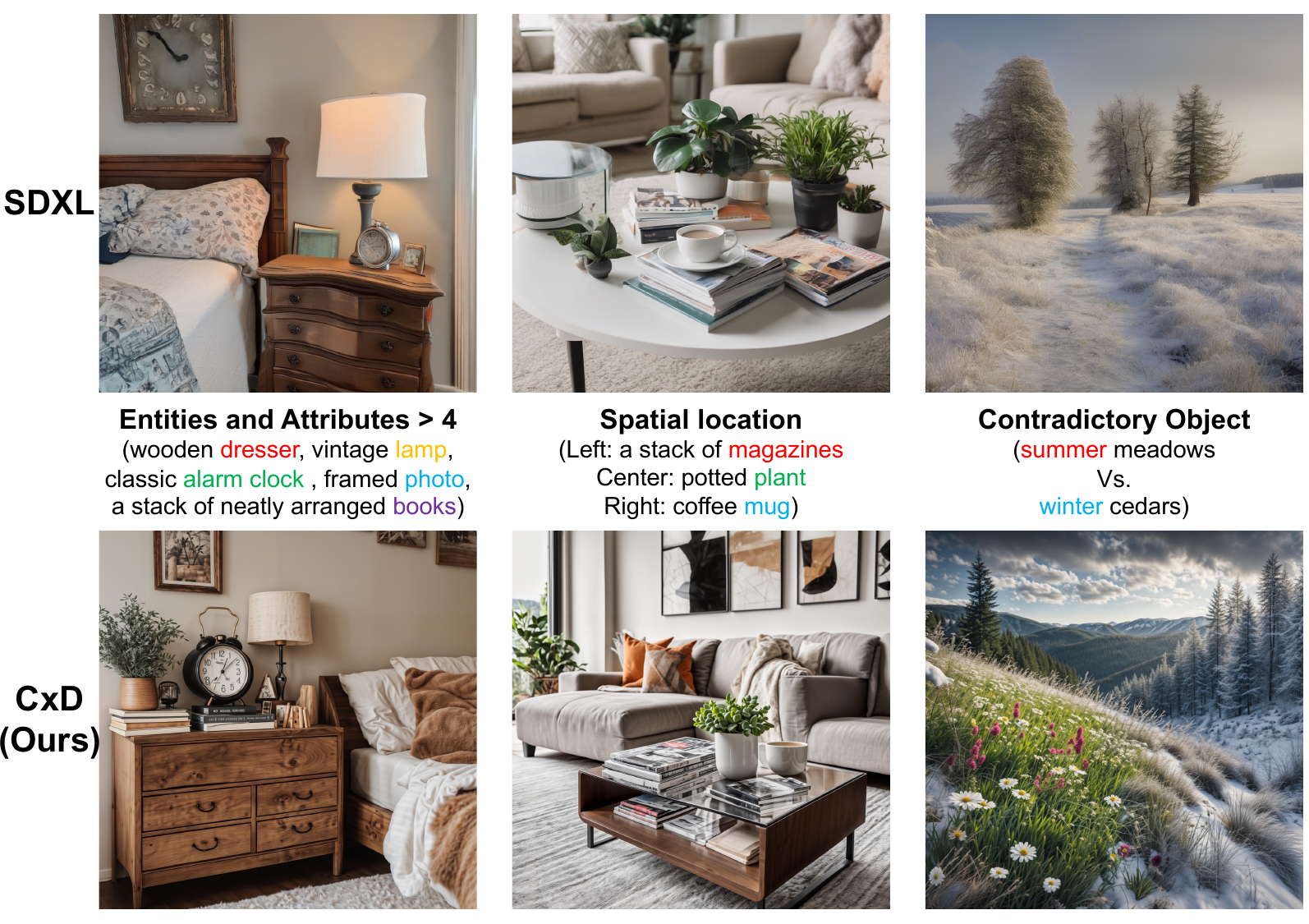}
    \caption{Limitations of pre-trained diffusion models in complex scene generation.}
    \label{resultxl}
\end{figure}

Recently, diffusion models \cite{sohl2015deep,song2020score,yang2023diffusion,jahn2021high} have represented a significant advancement in text-to-image generation, showcasing impressive capabilities in producing high-quality images from textual descriptions. However, despite their remarkable performance, these models face substantial challenges when tasked with generating complex scenes. Specifically, as illustrated in Figure \ref{resultxl}, when prompts involve multiple entities, intricate spatial positions, and conflicting relationships, the models often encounter issues such as entity omissions, spatial inconsistencies, and overall disharmony in the generated images.

Several works \cite{li2023gligen,xie2023boxdiff,qu2023layoutllm,chen2024training,yang2023law} have sought to address these challenges by incorporating additional layouts or boxes to decompose complex scene relationships. For instance, LAW-diffusion \cite{yang2023law} integrates layout configurations into the synthesis process to enhance the coherence of object relationships in complex scenes. Similarly, some approaches \cite{kim2023dense,wang2024compositional,yang2024mastering} utilize prompt-aware attention guidance to improve compositional text-to-image synthesis. Dense Diffusion \cite{kim2023dense}, for example, adjusts intermediate attention maps based on layout conditions to support detailed captions, while RPG \cite{yang2024mastering} employs MLLM as a global planner to decompose complex images into simpler tasks within subregions, introducing complementary regional diffusion for region-specific compositional generation. Despite these advancements, gaps remain when dealing with highly complex scene prompts, and the definition of a `complex scene' continues to be somewhat ambiguous.

In parallel, traditional art processes for creating complex scenes involve a meticulous three-stage approach: composition, painting, and retouching \cite{arnheim1954art,dow2023composition,gombrich1995story,schaeffer2023art}. Artists begin by sketching the overall layout and positioning of elements (composition), followed by detailed painting where the main features are developed (painting), and finally, they refine the artwork by adding details and correcting imperfections (retouching). This methodical approach ensures that every aspect of the scene is carefully considered and harmonized. Adapting this artistic process to model-based generation might provide a solution for complex scene generation.

To address these issues, we propose a novel training-free diffusion framework called \textbf{C}omple\textbf{x} \textbf{D}iffusion (\textbf{CxD}), which draws inspiration from the artistic creation process and divides the scene generation into three stages: composition, painting, and retouching. We start by leveraging the powerful reasoning abilities of chain-of-thought in a Large Language Model (LLM) to decompose and compose complex prompts according to \textbf{C}omplex \textbf{D}ecomposition \textbf{C}riteria (\textbf{CDC}). Next, CxD adapts the cross-attention layer to handle the simplified prompts and compositions derived from the LLM decomposition. Finally, we use a retouching model (ControlNet-tile\cite{zhang2023adding}) to enhance the details of complex scenes based on the attributes obtained during the LLM decomposition process. Comprehensive experiments across multiple tasks demonstrate that our proposed method consistently outperforms previous SOTA approaches, achieving significant improvements in generating high-quality, semantically consistent, and visually diverse images for complex scenes, even when conditioned on intricate textual prompts. In summary, our contributions can be outlined as follows:

\begin{itemize}
    \item \textbf{Definition and Criteria}. We provide a clear experimental definition of complex scenes and introduce \textbf{C}omplexity \textbf{D}ecomposition \textbf{C}riteria (\textbf{CDC}) to effectively manage complex prompts.
    \item \textbf{CxD Framework}: Drawing inspiration from the artistic creation process, we propose a training-free \textbf{C}omple\textbf{x} \textbf{D}iffusion (\textbf{CxD}) framework that divides the generation of complex scene images into three stages: composition, painting, and retouching.
    \item \textbf{Validation and Performance}. Extensive experiments demonstrate that CxD generates high-quality, consistent, and diverse images of complex scenes, even when dealing with intricate prompts
\end{itemize}

\section{Related Work}
\subsection{Complex Sense Generation}
Text-to-image(T2I) generation has made significant strides, but generating complex scenes remains challenging. Transformer-based models \cite{yang2022modeling,jahn2021high} have improved spatial layout modeling and data efficiency, enhancing texture, structure, and relational accuracy. GAN-based models \cite{goodfellow2014generative} leverage a generator-discriminator framework to produce images aligned with textual descriptions, but complex scenes with multiple object categories demand a highly proficient discriminator \cite{lee2023improving,hua2021exploiting,lee2023multi}. Recently, diffusion models \cite{dhariwal2021diffusion,ho2020denoising,rombach2022high} have gained popularity over GANs for their high-quality, diverse outputs. Notably, LAW-Diffusion \cite{yang2023law}  enhances object relationship coherence in complex scenes by integrating layout configurations into the synthesis process.

\subsection{Compositional Diffusion Generation}
Composition is vital in painting, and recent studies have explored compositional diffusion models to aid this process\cite{li2023gligen,yang2023reco,huang2023composer}. Some methods, like ControlNet \cite{zhang2023adding} and T2I-Adapter \cite{mou2024t2i}, enhance controllability by training additional modules, but this adds extra costs. Others, like Composable Diffusion \cite{liu2022compositional} MultiDiffusion \cite{bar2023multidiffusion}, and Dense Diffusion \cite{kim2023dense}, manipulate latent or cross-attention maps to control the model without extra training, often using bounding boxes to guide composition \cite{chen2024training,xie2023boxdiff}. However, these approaches struggle with complex prompts, leading to incomplete or distorted images. To address this, we propose an efficient, training-free method that maintains image controllability for complex prompts without additional costs.

\subsection{LLM for Image generation}
Large language models (LLMs) have revolutionized AI research, exemplified by models like ChatGPT \cite{wu2023brief} with strong language comprehension and reasoning. Leveraging LLMs, diffusion models improve text-image alignment and image quality \cite{wu2024self,yang2023reco,pan2023kosmos}. LLMs can control layout by assigning locations based on prompts \cite{cho2024visual,lian2023llm}, and models like LayoutGPT \cite{feng2024layoutgpt} enhance this by providing retrieval samples. Ranni \cite{feng2024ranni} adds a semantic panel with multiple attributes, while RPG \cite{yang2024mastering} uses LLMs for image composition planning. However, most methods rely on LLMs' inherent abilities or simple prompts. In contrast, we introduce complex decomposition criteria (CDC) to guide LLMs in helping diffusion models understand complex text prompts.
\begin{figure}
    \centering
    \includegraphics[width=\linewidth]{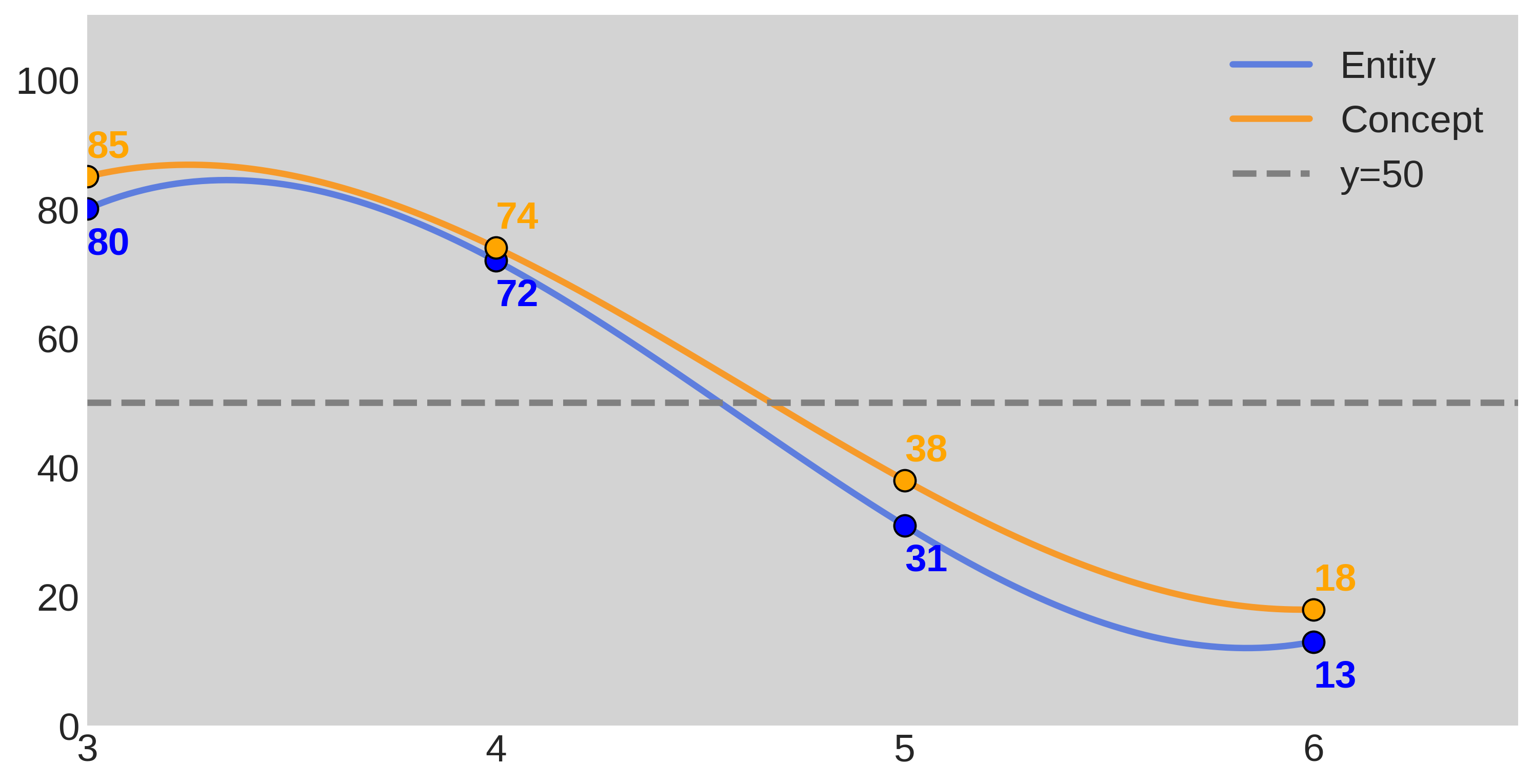}
    \caption{Performance trends of the SD XL model with varying numbers of entities and concepts}
    \label{line}
\end{figure}

\section{Complex Scene in Pre-Trained Diffusion}

As AI-generated content (AIGC) progresses, image generation, particularly complex scene generation, has become a significant focus and research challenge \cite{krishna2017visual,huang2023t2i}. Many studies\cite{yang2023law,yang2024mastering} describe complex scenes using vague terms like “multiple”, “different”, and “diverse”, without clearly defining what constitutes “complexity.” This ambiguity can lead to biased evaluations and ineffective solutions. Thus, a precise definition of “complexity” in scene generation is crucial.
\begin{figure*}[h]
    \centering
    \includegraphics[width=\linewidth]{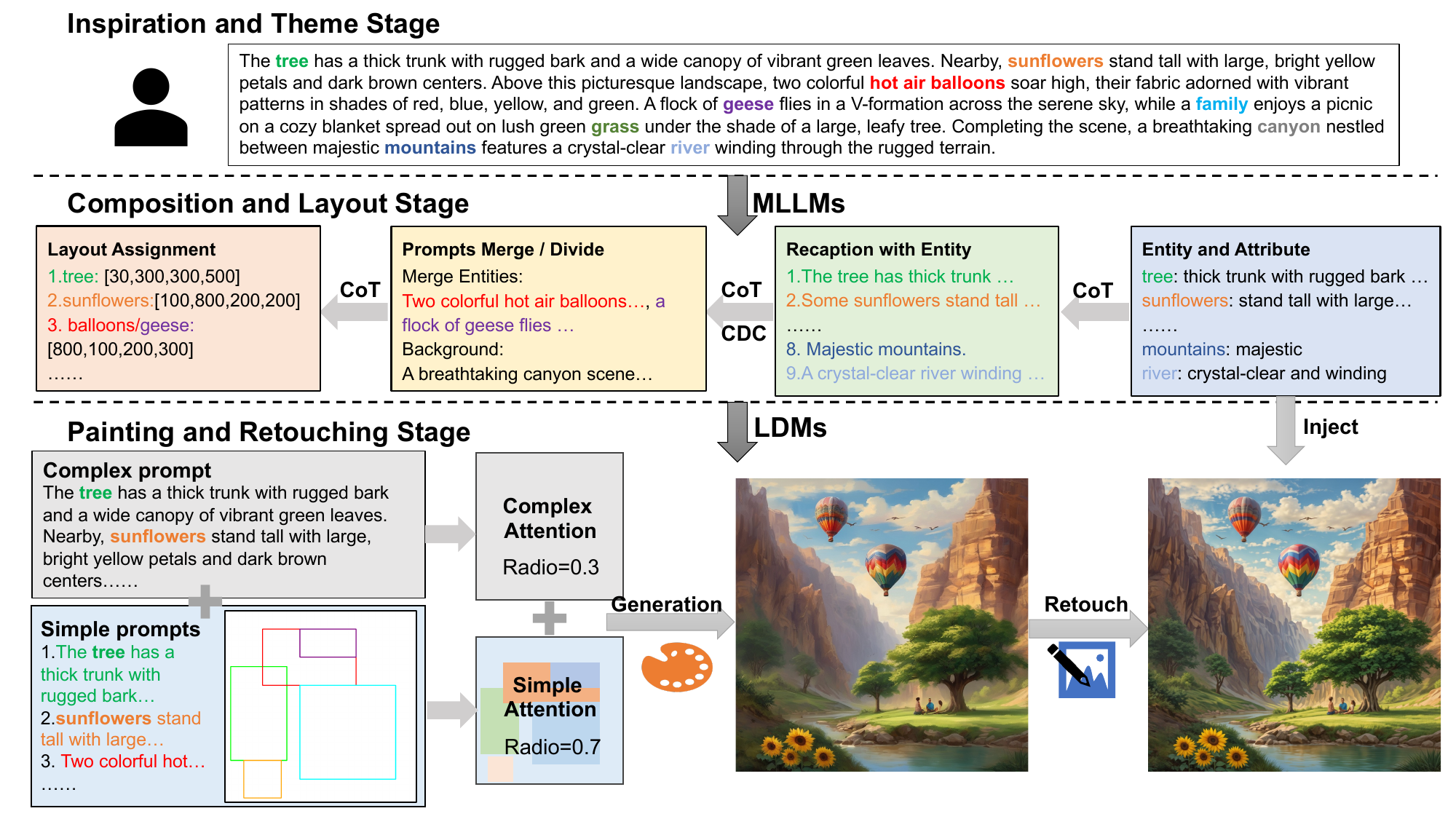}
    \caption{Overview of CxD framework for complex scene image generation.}
    \label{pipeline}
\end{figure*}
\subsection{Definition of `Complex' in Pre-Trained Diffusion}
Although some studies attempt to define “complexity” in terms of the number of entities, their attributes, and relationships \cite{krishna2017visual,huang2023t2i,yang2023law,yang2024mastering}, our experiments suggest these factors must be considered collectively. Generating multiple entities alone is not as challenging for diffusion models as combining these entities with specific attributes and relationships, which often results in omissions, inconsistencies, and disharmony.
 
Building on previous research\cite{huang2023t2i}, we redefine complexity in scene generation by considering four key factors: the number of entities, attributes, spatial positioning, and relationships among entities. Notably, relationships here refer to associations or conflicts rather than spatial arrangements. Our findings indicate that prompts containing conflicting entities are particularly challenging for diffusion models, often leading to inconsistent and visually unsatisfactory results, as shown in Figure \ref{resultxl}.

First, we tested diffusion models with prompts containing three to six positively correlated entities. As shown in Figure 2, when the number of entities reached five, nearly 70\% of the images showed omissions or disharmony, worsening with six entities. Next, we examined the effect of adding attributes. Next, we examined the impact of adding attributes. When the total number of entities and attributes was four, the results were similar to those with four entities alone, achieving a 74\% success rate. However, when the combined number increased to five, the success rate dropped to 38\%, indicating that the model struggles with prompts involving more than four concepts. (entities plus attributes). Lastly, we explored spatial positioning and relationships. Even with two entities, accurately capturing spatial relationships proved difficult, with success rates falling below 50\%. For conflicting relationships (e.g., desert vs. rainforest and summer meadows vs. winter ceders), the success rate plummeted to 10\%.

In summary, scenes with fewer than five concepts and no spatial or conflicting relationships are simple, while those exceeding these criteria are considered complex.

\subsection{Complex scene Decomposition Criteria (CDC)}
Based on our definition, we propose  \textbf{C}omplex Scene Decomposition Criteria (\textbf{CDC}) to help both humans and LLMs simplify complex prompts:

\begin{itemize}
    \item \textbf{Identify Conflicting Entities:} If conflicting entities are present, classify the scene as complex and separate these entities into different prompts.
    \item \textbf{Check for Spatial Relationships:} If spatial relationships are involved, classify the scene as complex and split entities with positional relationships into different prompts, maintaining their spatial context.
    \item \textbf{Evaluate Number of Concepts:} If a prompt contains more than four concepts, decompose it by entities, ensuring each prompt retains as many attributes as possible without exceeding four per entity.
\end{itemize}

This approach ensures that complex prompts are effectively simplified for more accurate and consistent image generation.

\begin{figure}
    \centering
    \includegraphics[width=\linewidth]{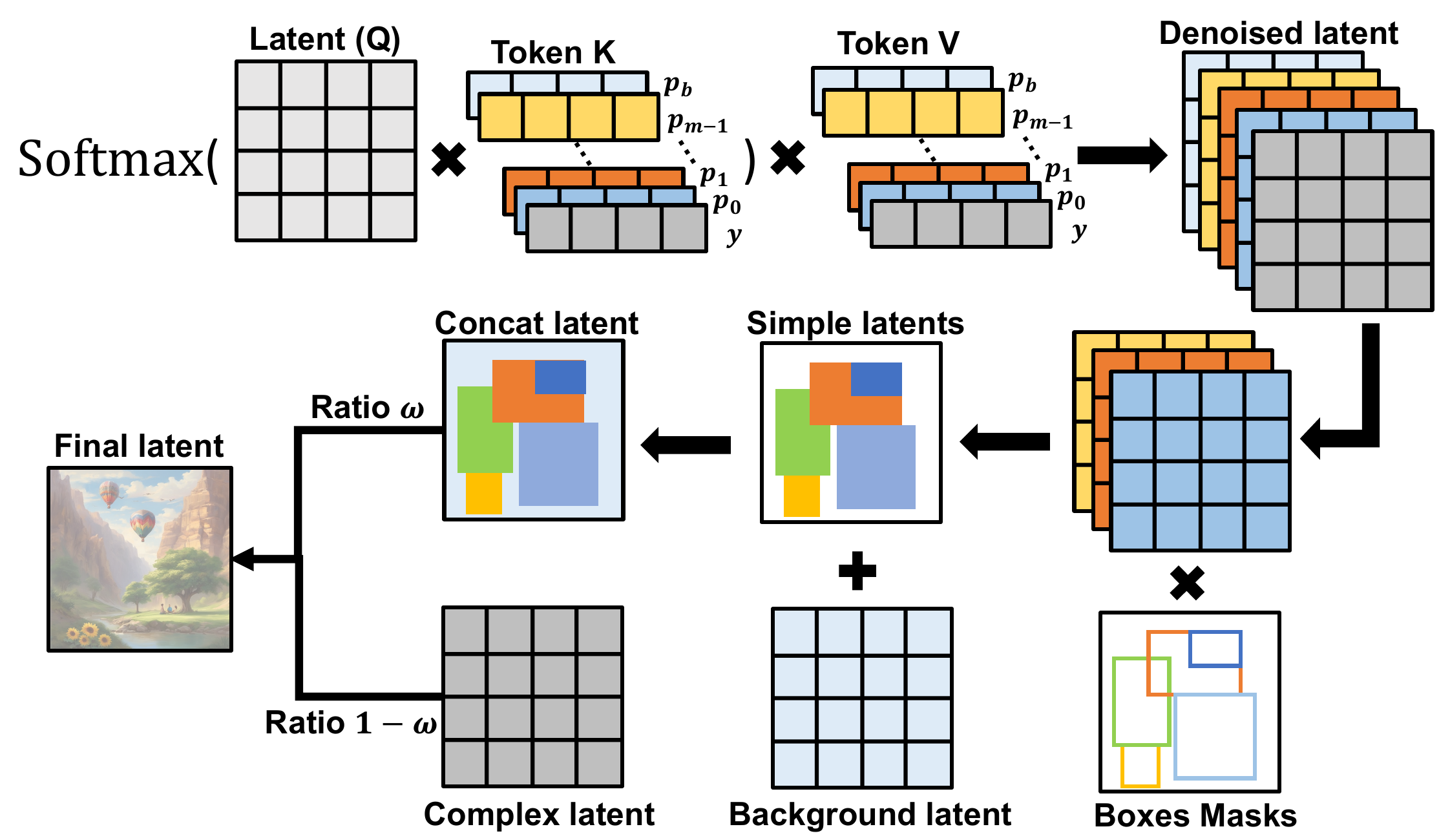}
    \caption{The demonstration of each sampling step in CxD.}
    \label{attention}
\end{figure}

\section{Method:CxD}

In this section, we present our training-free framework, \textbf{CxD}, which mirrors the artist’s drawing process by dividing complex scene generation into three stages: Composition, Painting, and Retouching, as shown in Figure \ref{pipeline}. Starting with a complex scene prompt, we use the Chain-of-Thought (CoT) approach in Large Language Models (LLMs) for composition. The LLM extracts entities and attributes, rephrases with entities, merges them, divides the background based on Complex Decomposition Criteria (CDC), and assigns layouts. CxD then computes and combines complex and simple cross-attention maps at each sampling step. Finally, attributes extracted by the LLM are injected into ControlNet tile \cite{zhang2023adding} for detailed retouching. Below, we detail the methods used in these three stages.

\subsection{Composition and layout generation with LLMs}
\subsubsection{Entities extraction}
Upon receiving a complex scene prompt $y$ from the user, we leverage the advanced language understanding and reasoning capabilities of the LLM to extract the entities $E$ and corresponding attributes $A$ from the prompt. This process can be described as follow:
\begin{equation}
    \{E_i\}_{i=0}^n={E_0,E_1,…,E_n}=LLM_{extract\_enti}(y)
\end{equation}
\begin{equation}
\begin{split}
        \{A_i\}_{i=0}^n=
       &\ \{A_0,A_1,...,A_n\}= \{(a_0^0,...,a_0^j),...,
         (a_0^n,...,a_n^k)\}\\
   &\ =LLM_{extract_attr}(y)
\end{split}
\end{equation}
where $n$ denotes the number of entities in the complex scene prompt, with $E_i$ representing the $i$-th entity. $A_i$ denotes the set of attributes corresponding $E_i$, and $a_i^j$ refers to the $j$-th attribute of the $i$-th entity. It is important to note that the number of attributes for different entities is not necessarily equal, so $j$ may differ from $k$. Furthermore, the number of entities $n$ and the attributes for each entity $j$ are not predefined hyperparameters but are determined dynamically through LLM heuristics.
\subsubsection{Prompts recaption}
Inspired by RPG \cite{yang2024mastering}, which utilizes LLMs to recaption prompts and plan for region divisions with chain-of-thought (CoT). We also employ LLM to recaption prompt into sub-prompts based on the extracted entities $E$ and corresponding attributes $A$. These sub-prompts are designed to be as consistent as possible with the relevant description in the original complex prompt. This process can be denoted as:
\begin{equation}
    \{\hat{y}_i
    \}_{i=0}^n=\{\hat{y}_0,…,\hat{y}_n \}=LLM_{recaption}(E_i,A_i,y)
\end{equation}
where $\hat{y}_i$ represents the sub-prompt describing the $i$-th entity, with each entity corresponding to a distinct sub-prompt.
\subsubsection{Prompts merge or divide}
After recaptioning, the sub-prompts have been simplified a lot compared to the original complex prompt. However, we cannot guarantee that all sub-prompts will be sufficiently simple for the generative model, as some may still be relatively complex. In addition, some sub-prompts may be very simple on their own, even when combined, the overall prompt might still be straightforward for the generative model. To ensure the quality and efficiency of image generation, we use LLM to merge or split sub-prompts based on the Complex Decomposition Criteria (CDC). The results of merging or splitting are documented as simple prompts:
\begin{equation}
    \{p_i\}_{i=0}^m=\{p_0,p_1,...,p_m\}=LLM_{md}(\hat{y})
\end{equation}
where $m$ is the number of simple prompts, $m\leq n$. $LLM_{md}$ denotes the operation of merging and dividing prompt.

For all simple prompts, we instruct the LLM to filter out background prompt $p_b$, which is image background and do not participate in layout assignment.
\renewcommand{\dblfloatpagefraction}{.9}
\begin{figure*}[h]
    \centering
    \includegraphics[width=\linewidth]{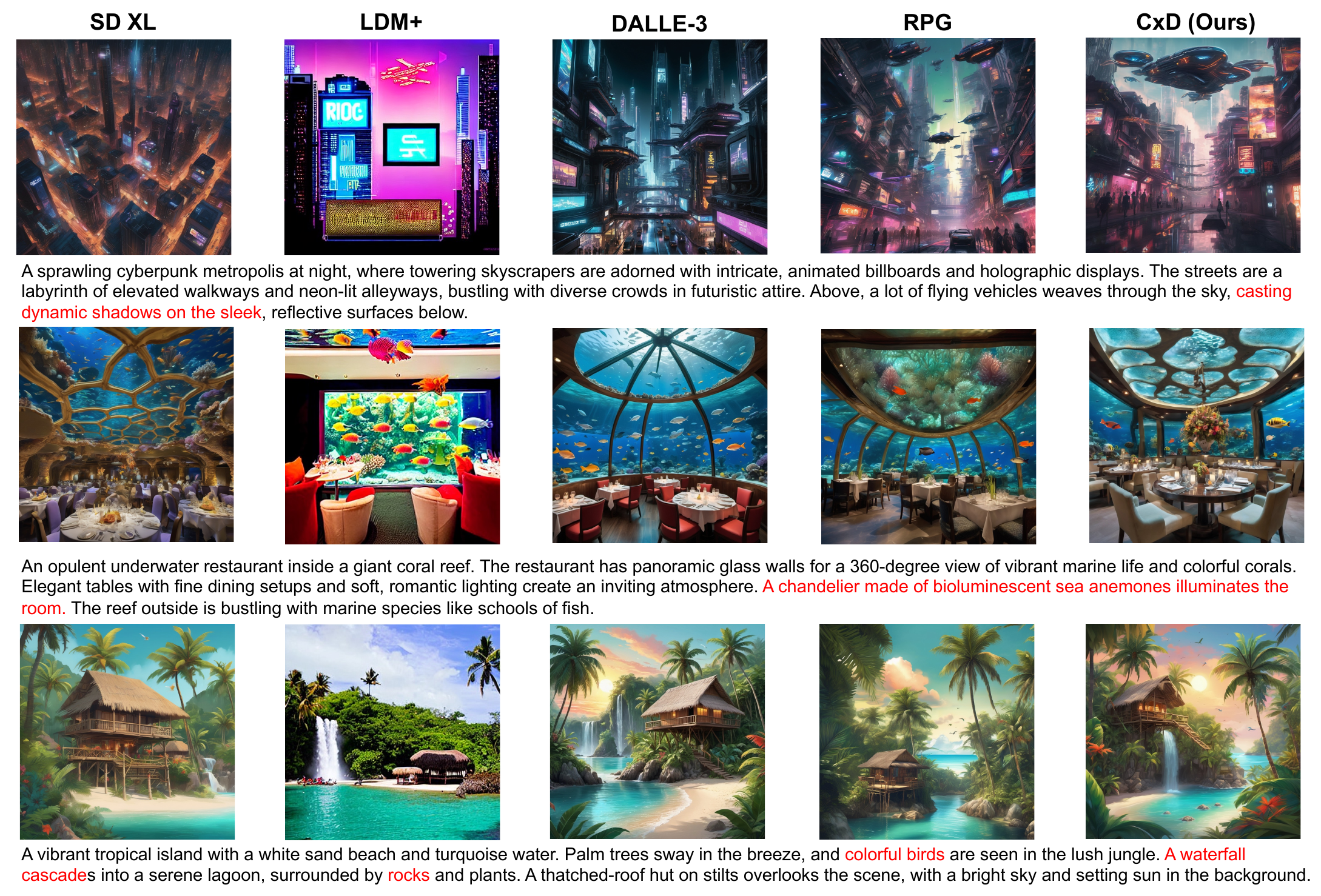}
    \caption{Qualitative comparison between CxD and SOTA text-to-image models}
    \label{compare}
\end{figure*}
\subsubsection{Layout assignment}
Except for the background prompt, all simple prompts are assigned layouts by the LLM to complete the final composition. These layouts are specified with coordinates in the $(x, y, width, height)$ format. Specifically, the LLM prioritizes the positional and size relationships between different entities mentioned in the prompt to assign bounding boxes. For entities without specified relationships, the LLM uses its common sense knowledge to determine appropriate bounding boxes. The bounding boxes are denoted as $\{B_i\}_{i=0}^{m-1}=\{B_0,B_1,…,B_{m-1}\}$. The area not covered by these layouts is reserved for the background prompt $\{BG\}$.

Finally, we sort the layouts assigned by the LLM in descending order of their area size and adjust the order of the corresponding simple prompts accordingly. This approach aligns with the artist’s practice of prioritizing the main subject first and also helps prevent smaller objects from being obscured by larger ones when entities overlap during image generation.

\subsection{Cross-Attention Modulation}
As analyzed in the previous section, diffusion models tend to be less effective with complex scenarios involving more than four concepts. To address this challenge, we modulate the cross-attention to adapt to composition generated by LLM, facilitating efficient handling of complex scene prompts, as shown in Figure \ref{attention}.

\subsubsection{Prompt batch process}
Since the complex scene prompt is decomposed into various simple prompts by the LLM, we sample the same latent $z_t$ as the query at each timestep to ensure image consistency. Additionally, we construct a prompt batch consisting of the complex prompt y, simple prompts $\{p_i\}_{i=0}^{m-1}$ and background prompt $p_b$. At each timestep, this prompt batch is fed into the denoising network, which manipulates the cross-attention layers to generated different latents in parallel: the complex latent $z^c$, simple latents $\{z^i\}_{i=0}^{m-1}$ and background latent $z^b$. The process can be formulated as follows:
\begin{equation}
    z_{t-1}^i=Softmax(\frac{(W_Q\cdot\phi(z_t)(W_K\cdot\psi(p_i))}{\sqrt{d}})(W_V\cdot\psi(p_i))
\end{equation}
where $W_Q$, $W_K$, $W_V$ are linear projections, and $d$ is the latent projection dimension of the keys and queries. $\phi(z_t)$ and $\psi(p_i )$ denote the embeddings of the latent $z_t$ and the simple prompt $p_i$ at the $t$-th timestep, respectively.
\renewcommand{\dblfloatpagefraction}{.9}
\renewcommand{\arraystretch}{1.2} 
\begin{figure*}[h]

    \centering
    \includegraphics[width=\linewidth]{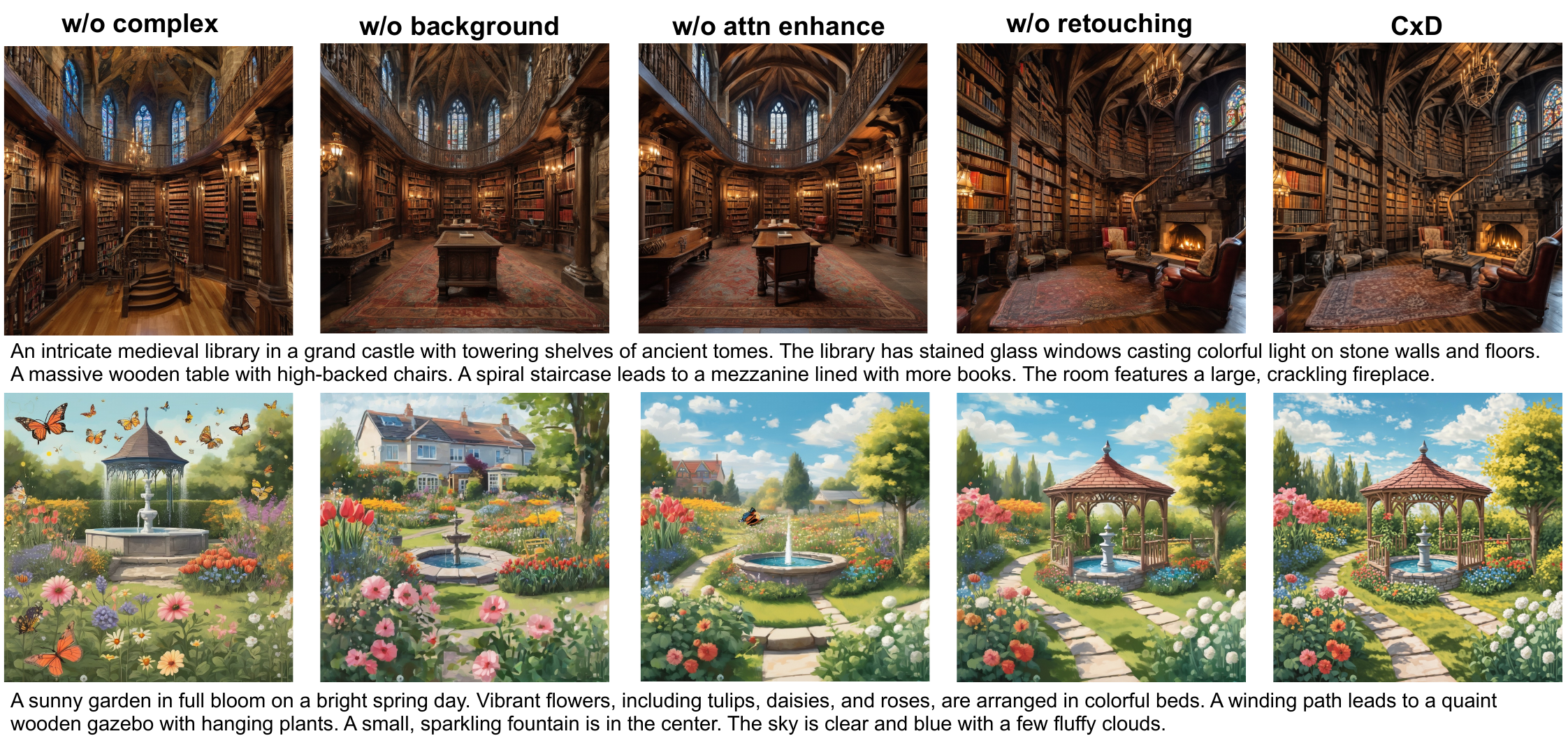}
    \caption{Visualization of CxD ablation study}
    \label{ablation}
\end{figure*}
\subsubsection{Cross-attention enhancement modulation}
To avoid missing concepts involving numerous entities and attributes and to enhance details, we propose attention enhancement modulation. Specifically, we resize the bounding boxes generated by the LLM to match the shape of the latent $z_{t-1}^i$ as follows:
\begin{equation}
    \hat{B}=Resize(B_i,l_{scale})
\end{equation}
where $l_{scale}$ represents the scaling factor for the latent.

For each latent, we emphasize the region corresponding to bounding box $\hat{B}_i$ while minimizing the influence of the surrounding area. This process can be described as:
\begin{equation}
    z_{t-1}^i+=\lambda_{pos}\cdot M_{\hat{B}}^i\cdot(Max(z_{t-1}^i)-z_{t-1}^i)
\end{equation}
\begin{equation}
    z_{t-1}^i-=\lambda_{neg}\cdot (\sim M_{\hat{B}}^i)\cdot(z_{t-1}^i-Min(z_{t-1}^i))
\end{equation}
where $\lambda_{pos}$ and $\lambda_{neg}$ are hyperparameters that control the extent of modulation. $M_{\hat{B}}$ and $(\sim M_{\hat{B}}^i)$ are masks indicating the regions within and outside the bounding box $\hat{B}^i$, respectively. 

After modulate the results, we concatenate the results of all simple prompts’ denoising latents according to the area of the bounding boxes to achieve control over the positional relationship. The areas not covered by the bounding boxes are filled with the background denoising latent results. We define this process as:

\begin{equation}
    z_{t-1}^{cat}=Concat(\{z_{t-1}^i(\hat{B_i})\}_{i=0}^{m-1},\{z_{t-1}^b(\sim \hat{B})\})
\end{equation}
where $z_{t-1}^b$ denotes the background denoising latent and $(\sim \hat{B})$ represents region uncovered by the bounding boxes.

Finally, to ensure a seamless transition between regions and harmonious blending of the background with the entities, we blend the complex prompt latent $z_{t-1}^c$ with the concatenated latent $z_{t-1}^{cat}$ using a weighted sum. This approach integrates both elements effectively to produce the final denoised output.
\begin{equation}
    z_{t-1}=\omega\cdot z_{t-1}^{cat}+(1-\omega)\cdot z_{t-1}^c
\end{equation}
where $\omega$ is the weight used to balance the contributions of the complex prompt and the simple prompts.

To tackle the challenges of complex scenes, we decompose complex prompts into simpler ones to manage concept overload. Bounding boxes from the LLM help create precise latent representations for each simple prompt, ensuring accurate positional control. Generating each latent independently minimizes conflicts between entities. In summary, CxD effectively addresses the issues related to complex scenes.
\subsection{Retouching with ControlNet tile Model}
Our method effectively generates images that align with the descriptions of complex prompts. However, when the number of entities and attributes exceeds the capacity of the pre-trained diffusion model, some local details unrelated to the complex prompt may be lost or blurred. To address this, we employ retouching models to refine the results, much like an artist adding finishing touches to a painting. We supply the entities and attributes extracted by the LLM as details to a ControlNet \cite{zhang2023adding} extension—ControlNet-tile model—which enhances the image by correcting defects and incorporating new details. After applying ControlNet-tile, the image retains its original semantics but gains enhanced clarity in details and textures. So far, we have completed the creation of a complex scene image through three stages—composition, painting, and retouching—much like the process an artist would follow.
\renewcommand{\dblfloatpagefraction}{.9}
\begin{table*}[h]
\centering
\caption{Evaluation results on T2I-CompBench. We denote the best score in blue, and the second-best score in green.}
\begin{tabular}{lcccccc}

\hline
\multirow{2}{*}{Model} & \multicolumn{3}{c}{Attribute Binding} & \multicolumn{2}{c}{Object Relationship} & \multirow{2}{*}{Complex} \\
\cline{2-4} \cline{5-6}
 & \multicolumn{1}{c}{Color} & \multicolumn{1}{c}{shape} & \multicolumn{1}{c}{Texture} & \multicolumn{1}{c}{Spatial} & \multicolumn{1}{c}{Non-Spatial} & \\

\hline
Stable Diffusion V1.4\cite{rombach2022high} & 0.3765& 0.3576 &0.4156& 0.1246& 0.3079 &0.3080\\
Stable Diffusion v2 \cite{rombach2022high}& 0.5065& 0.4221& 0.4922 &0.1342 &0.3096 &0.3386\\
 Composable Diffusion\cite{liu2022compositional} & 0.4063& 0.3299 &0.3645 &0.0800 &0.2980 &0.2898\\ Structured Diffusion\cite{feng2022training} & 0.4990& 0.4218 &0.4900 &0.1386 &0.3111& 0.3355\\ Attn-Exct v2\cite{chefer2023attend} & 0.6400 &0.4517 &0.5963 &0.1455 &0.3109 &0.3401 \\
 GORS\cite{huang2023t2i} & 0.6603 &0.4785& 0.6287& 0.1815 &0.3193 &0.3328 \\
 DALL-E 2\cite{ramesh2022hierarchical} & 0.5750& 0.5464 &0.6374 &0.1283& 0.3043 &0.3696\\
 SDXL\cite{podell2023sdxl} & 0.6369& 0.5408 &0.5637 &0.2032 &0.3110 &0.4091\\
 PixArt-$\alpha$ \cite{chen2023pixart} & 0.6886 &0.5582 &0.7044& 0.2082& 0.3179& 0.4117\\
 ConPreDiff\cite{yang2024improving} & 0.7019 &0.5637 &0.7021 &0.2362 &0.3195 &0.4184\\
 RPG\cite{yang2024mastering} & \cellcolor[HTML]{E9F9E3}0.8335 &\cellcolor[HTML]{D9EAF4}0.6801 &\cellcolor[HTML]{E9F9E3}0.8129  &\cellcolor[HTML]{E9F9E3}0.4547  &\cellcolor[HTML]{E9F9E3}0.3462 &\cellcolor[HTML]{E9F9E3}0.5408\\
 \hline
 CxD(Ours)&\cellcolor[HTML]{D9EAF4}0.8562&\cellcolor[HTML]{E9F9E3}0.6533&\cellcolor[HTML]{D9EAF4}0.8563&\cellcolor[HTML]{D9EAF4}0.6241&\cellcolor[HTML]{D9EAF4}0.5426&\cellcolor[HTML]{D9EAF4}0.6713\\
\hline
\label{table1}
\end{tabular}
\end{table*}

\section{Experiment}
\subsection{Experiment Setting}
For our CxD framework, we utilize the open-source LLaMA-2 \cite{touvron2023llama} 13B version as our large language model (LLM) and the Stable Diffusion XL \cite{podell2023sdxl} version as our pre-trained diffusion model. However, CxD is designed to be a general and extensible framework, capable of integrating various LLM architectures. All experiments in this study were conducted on an NVIDIA RTX 3090 GPU. Generating a complex scene image with CxD takes approximately 2 minutes, including the time required for processing complex prompts with the LLM. We have carefully crafted task-aware templates and high-quality in-context examples to leverage the chain-of-thought (CoT) capabilities of the LLM effectively.

\subsection{Qualitative Assessment}
We evaluated CxD’s performance against various complexity indicators, including the number of concepts, spatial locations, and conflicting relationships. Figure \ref{resultxl} compares results from the SD XL \cite{podell2023sdxl} model and CxD. The top row shows SD XL struggling with high complexity, including distortion and inaccuracies in spatial positioning when handling prompts with five entities and attributes. It also tends to ignore one side of entity conflicts. In contrast, CxD effectively manages high complexity, precise spatial arrangements, and conflicting entities, producing consistently harmonious and visually appealing images.

We compared CxD with previous state-of-the-art text-to-image models, including SDXL \cite{podell2023sdxl}, LDM+ \cite{lian2023llm}, DALLE-3 \cite{lee2023improving}, and RPG \cite{yang2024mastering}. LDM+ and RPG use LLMs for composition assistance. As shown in Figure \ref{compare}, SDXL and LDM+ struggle with complex prompts, resulting in images that do not fully meet prompt expectations. While DALLE-3 and RPG effectively capture overall content, they sometimes miss local details in complex prompts (e.g., the red part in Figure \ref{compare}). In contrast, CxD decomposes complex prompt into simple prompts, ensuring no entities or attributes are omitted. Consequently, CxD excels in managing both overall semantics and local details, demonstrating its effectiveness in handling complex scenes.

\subsection{Quantitative Experiments}
We compared our CxD with previous SOTA text-to-image models using the T2I-Compbench benchmark \cite{huang2023t2i}. As shown in Table \ref{table1}, Our CxD consistently outperforms all others in both general text-to-image generation and complex generation, with RPG coming in second. This highlights the superiority of our approach in handling complex scene generation tasks.
We evaluated our CxD model against previous SOTA text-to-image models using the T2I-Compbench benchmark \cite{huang2023t2i}. As shown in Table \ref{table1}, our model sets a new SOTA benchmark in most tasks, particularly excelling in object relationships and complex scenarios, and significantly outperforms the second-best method. This exceptional performance can be attributed to the strong alignment of these tasks with our proposed Complex Decomposition Criteria (CDC), demonstrating the superiority of our approach in addressing complex scene generation.

\subsection{Ablation Study}
We evaluate each component of our CxD framework: (a) Complex prompt latent, (b) Background prompt latent, (c) Attention enhancement modulation, and (d) Image retouching, as shown in Figure \ref{ablation}. The first column shows images without complex prompt latent, resulting in disjointed and inconsistent outputs. The second column, lacking background prompt latent, displays backgrounds that do not meet prompt requirements. The third column, without attention enhancement modulation, results in obscured entities. The fourth column, missing modification, produces images with blurred details due to too many entities. The final column shows our CxD framework’s output, preserving semantics and enhancing details, highlighting the importance of each CxD component in generating complex scenes.
\section{Conclusion}
In this paper, we propose CxD, a training-free diffusion framework designed to tackle the challenges of complex scene generation. We define ‘complex scenes’ with precision and provide a set of Complex Decomposition Criteria (CDC) for both humans and large language models (LLMs) to effectively handle complex scene prompts. The CxD framework divides the generation process into three stages—composition, painting, and retouching—mirroring the traditional artist’s approach to drawing. Our experimental results demonstrate that CxD performs well in generating complex scenes. Future work will focus on integrating additional modal data as input conditions to further enhance controllability.

\bibliographystyle{plain}
\bibliography{ijcai19}

\end{document}